\documentclass[letterpaper, 10 pt, conference]{ieeeconf}  
\IEEEoverridecommandlockouts                              
\usepackage{graphicx}
\usepackage{cite}
\usepackage{caption}
\usepackage{subfigure}
\usepackage{hyperref}
\usepackage{multirow}
\usepackage{wrapfig}
\usepackage{amssymb}
\usepackage{algorithm}
\usepackage{algorithmic}
\usepackage{url}

\title{\LARGE \bf High precision grasp pose detection in dense clutter* \thanks{* The first two authors contributed equally to the paper.}}
\author{Marcus Gualtieri$^\dagger$, Andreas ten Pas$^\dagger$, Kate Saenko$^\ddagger$, Robert Platt$^\dagger$ \\
$\left.^\dagger \right.$College of Computer and Information Science, Northeastern University \\
$\left.^\ddagger \right.$Department of Computer Science, University of Massachusetts, Lowell}
\begin{document}

\maketitle

\thispagestyle{empty}
\pagestyle{empty}

\begin{abstract}

This paper considers the problem of grasp pose detection in point clouds. We follow a general algorithmic structure that first generates a large set of 6-DOF grasp candidates and then classifies each of them as a good or a bad grasp. Our focus in this paper is on improving the second step by using depth sensor scans from large online datasets to train a convolutional neural network. We propose two new representations of grasp candidates, and we quantify the effect of using prior knowledge of two forms: instance or category knowledge of the object to be grasped, and pretraining the network on simulated depth data obtained from idealized CAD models. Our analysis shows that a more informative grasp candidate representation as well as pretraining and prior knowledge significantly improve grasp detection. We evaluate our approach on a Baxter Research Robot and demonstrate an average grasp success rate of 93\% in dense clutter. This is a 20\% improvement compared to our prior work.


\end{abstract}


\section{Introduction}

Grasp pose detection is a relatively new approach to perception for robot grasping. Most approaches to grasp perception work by fitting a CAD model of the object to be grasped to sensor data (typically a point cloud). Grasp configurations calculated for the CAD model are thereby transformed into real world robot configurations. Unfortunately, this approach inherently makes a closed world assumption: that an accurate CAD model exists for every object that is to be grasped. Moreover, registering a CAD model to a partial and incomplete point cloud accurately and robustly can be very challenging. In contrast, grasp pose detection (GPD) characterizes the local geometry and/or appearance of graspable object surfaces using machine learning methods. Because GPD methods detect grasps independently of object identity, they typically generalize grasp knowledge to new objects as well.

Unfortunately, GPD methods have not yet been demonstrated to be reliable enough to be used widely. Many GPD approaches achieve grasp success rates (grasp successes as a fraction of the total number of grasp attempts) between 75\% and 95\% for novel objects presented in isolation or in light clutter. Not only are these success rates too low for practical grasping applications, but the light clutter scenarios that are evaluated often do not reflect the challenges of real world grasping.

In this paper, we evaluate a series of ideas intended to improve GPD performance using data derived from the BigBird dataset~\cite{Singh2014} -- an online dataset comprised of depth scans and RGB images of real objects. First, we focus on representation: we propose two new ways of representing the grasp to the learning system and compare them to the approaches proposed in~\cite{Kappler2015} and~\cite{Herzog2012}. Second, we evaluate the advantages of using prior category and/or instance knowledge of the object being grasped to improve grasp detection accuracy. One would expect that even general knowledge of the type of object that is being grasped should improve grasp detection accuracy. Here, we quantify this effect. Third, we evaluate the advantages of using training data derived from idealized CAD models to help pretrain the convolutional neural network used to do grasp classification. Although differences between simulated depth scans and real scans make it difficult to learn using simulated data alone, we have found that pretraining on the simulated (CAD model) data and finetuning on data from real scans is of benefit. Here, we quantify how much advantage pretraining on CAD models can confer. All of the above ideas are explored in the context of the general algorithmic structure used in~\cite{tenPas2015}, where a set of 6-DOF grasp candidates are first generated and then classified. We expect our conclusions to generalize fairly well to other approaches that follow the same algorithmic structure.

We also evaluate our approach on a Baxter Research Robot using an experimental protocol for grasping in dense clutter proposed in our prior work. First, objects are selected randomly and ``poured'' onto a tray in front of the robot. The robot must then grasp and remove as many objects as possible. Our results show that, out of 288 grasp attempts, our system failed to grasp only 20 times (a $_{\widetilde{~}}$93\% grasp success rate) in this scenario. This is a 20\% improvement over the success rate reported in our prior work~\cite{tenPas2015} \footnote{GPD code available at: \url{https://github.com/atenpas/gpd}}.

\section{Prior Work}

Grasp pose detection is distinguished from other approaches to robot grasping because it attempts to characterize graspable object surfaces in terms of local features rather than gross object geometry. For example, Lenz {\em et al.} model a graspable geometry as an oriented rectangle in an RGBD image~\cite{Lenz2015}. Given a number of candidate rectangles, machine learning methods trained on human-labeled data are used to predict which rectangles are grasps and which are not. An important characteristic of this work is that grasps are detected in the plane of the RGBD sensor: each detection corresponds to an $x, y, \theta$ position and orientation in an RGBD image. In order to use these detections to grasp, the gripper must approach the grasp target from a direction roughly orthogonal to the plane of the RGBD sensor. Several other approaches in the literature also detect grasps as an $x, y, \theta$ position and orientation in an RGBD image. For example, Pinto and Gupta take a similar approach except that the training data comes from on-line experience obtained by the robot during an automated experience-gathering phase~\cite{Pinto2016}. Using the same hand-labeled dataset, Redmon and Angelova pose grasp detection as a regression problem and solve it using convolutional neural network (CNN) methods~\cite{Redmon2015}.

A key limitation of detecting grasps as an $x, y, \theta$ pose in an RGBD image is that it constrains the robot hand to approach the object from one specific direction. This is a serious limitation because it is often easiest to grasp different objects in the same scene from different directions. Fischinger and Vincze take a step toward relaxing this constraint by detecting a grasp as an $x, y, \theta$ pose in a heightmap~\cite{Fischinger2012}. Since different heightmaps can be constructed from the same point cloud at different elevations, this enables the algorithm to control the grasp approach direction. The grasp template approach of Herzog {\em et al.} is still more flexible because it aligns the approach direction on a grasp-by-grasp basis with the object surface normal at the grasp  point~\cite{Herzog2012}. Kappler, Bohg, and Schaal show that the grasp templates proposed by Herzog {\em et al.} can be combined with a CNN-based grasp classifier~\cite{Kappler2015}. Finally, ten Pas and Platt propose a geometry-based method of generating grasp candidates and propose a representation that can be viewed as a variation on the template-based approach of Herzog~\cite{tenPas2015}.

An alternative approach is to demonstrate grasps on a set of objects to a robot and then to transfer these grasps to novel objects. While Kroemer {\em et al.} use actions afforded by object parts to learn the shape of the part~\cite{Kroemer2012}, Detry {\em et al.}~\cite{Detry2013} learn the geometry of typically grasped object parts. Kopicki {\em et al.} optimizes over the combination of a contact and a hand configuration model to generate grasp candidates~\cite{Kopicki2015}. In comparison, our approach does not require human demonstration.

\section{Grasp Pose Detection}

\subsection{Problem statement}

The input to our algorithm is a 3-D point cloud, $\mathcal{C} \subset \mathbb{R}^3$. Each point in the cloud is paired with at least one viewpoint ({\em i.e.} camera location) from which that point was observed, $V: \mathcal{C} \rightarrow \mathcal{V}$, where $\mathcal{V} \subset \mathbb{R}^3$ denotes the set of viewpoints. Let $CV = (\mathcal{C}, \mathcal{V}, V)$ denote the combined cloud and viewpoints. The algorithm also takes as input the geometric parameters of the robot gripper to be used, $\theta$, and a subset of points, $\mathcal{C_G} \subset \mathcal{C}$ , that identify which parts of the point cloud contain objects to be grasped. 

The output of the algorithm is a set of robot hand poses, $H \subset SE(3)$, such that if the robot hand is moved to a hand pose in $H$ and the gripper fingers are closed, then force closure is expected to be achieved for some object in the scene. This paper only considers the case where the robot hand is a 1-DOF two-fingered  parallel jaw gripper. Although the current version of the algorithm takes a point cloud as input, it would be easy to modify it to take a truncated signed distance function (TSDF) instead.

\subsection{Outline of the GPD algorithm}
\label{sect:gpd}

The grasp pose detection algorithm follows the two steps shown in Algorithm~\ref{alg:1}. 

\begin{algorithm}
\caption{GPD}
\vspace{0.05in}
{\bf Input:} a point cloud and associated viewpoints, $CV$; \\
a subset of the cloud where grasping is to occur, $\mathcal{C_G}$; \\
hand geometry, $\theta$ \\
{\bf Output:} a set of 6-DOF grasp configurations, $H \subset SE(3)$\\
\vspace{-0.15in}
\label{alg:overall}
\begin{algorithmic}[1]
\STATE $C = Sample\_Gasp\_Candidates(CV,\mathcal{C_G},\theta)$
\STATE $H = Classify(C,CV)$
\end{algorithmic}
\label{alg:1}
\end{algorithm}

\begin{figure}[b]
\begin{center}  
  \subfigure[]{\includegraphics[height=0.8in]{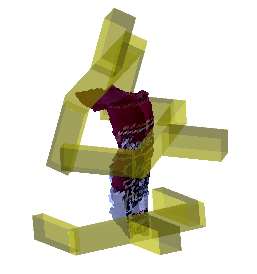}}
  \subfigure[]{\includegraphics[height=0.8in]{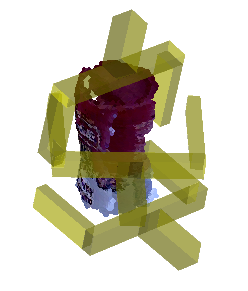}}
  \subfigure[]{\includegraphics[height=0.8in]{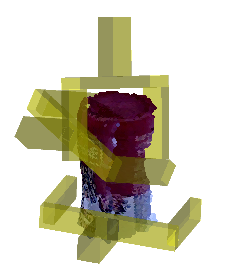}}
  \subfigure[]{\includegraphics[height=0.8in]{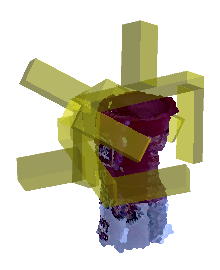}}
\end{center}
  \caption{Illustrations of grasp candidates found using our algorithm. Each image shows three examples of a gripper placed at randomly sampled grasp candidate configurations.}
  \label{fig:sampling}
\end{figure}

In Step 1, we sample several thousand grasp candidates. Each grasp candidate is a 6-DOF hand pose, $h \in SE(3)$. First, we sample points uniformly at random from $\mathcal{C_G}$. For each sample, we calculate a surface normal and an axis of major principle curvature of the object surface in the neighborhood of that point. Potential hand candidates are generated at regular orientations orthogonal to the curvature axis. At each orientation, we ``push'' the hand forward from a collision-free configuration until the fingers first make contact with the point cloud (a similar pushing strategy is used by Kappler { \em et al.} as part of the grasp candidate generation process~\cite{Kappler2015}). Once we have an in-contact hand configuration, we check to see whether any points from the cloud are contained between the fingers. If none are contained, then that grasp candidate is discarded. This process continues until a desired number of grasp candidates are generated. Figure~\ref{fig:sampling} illustrates candidates found using this method. Full details on the sampling method can be found in~\cite{tenPas2015}. 

In Step 2 of Algorithm~\ref{alg:1}, we classify each candidate as a grasp or not using a four-layer convolutional neural network (CNN). CNNs have become the standard choice for GPD classification and ranking~\cite{Kappler2016}. The main choice here is what CNN structure to use and how to encode to the CNN the geometry and appearance of the portion of the object to be grasped. We borrow the structure used by LeNet~\cite{LeCun1998} (the same structure is used by Kappler {\em et al.}): two convolutional/pooling layers followed by one inner product layer with a rectified linear unit at the output and one more inner product layer with a softmax on the output. The outputs, kernel size, pooling strides, {\em etc.} are all identical with those used by the LeNet solver provided in Caffe~\cite{Jia2014}. We used a learning rate of 0.00025 in all experiments. The next section discusses how we encode the geometry and appearance of the part of the object to be grasped.

\subsection{Grasp representation}
\label{sect:representation}

\begin{figure}[b]
\begin{center}  
  \subfigure[]{\includegraphics[height=1.2in]{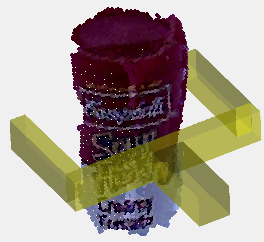}}
  \subfigure[]{\includegraphics[height=1.2in]{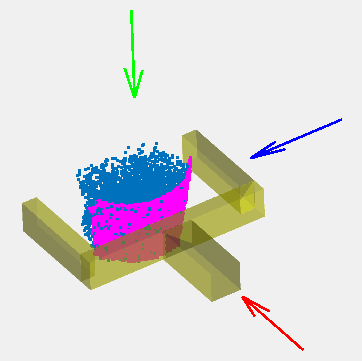}} 
  \\
  \subfigure[]{\includegraphics[height=0.8in]{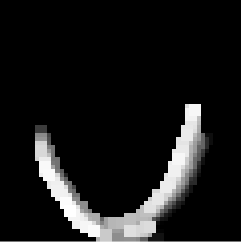}}
  \hspace{0.3in}
  \subfigure[]{\includegraphics[height=0.8in]{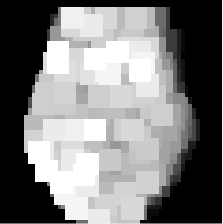}}
  \hspace{0.3in}
  \subfigure[]{\includegraphics[height=0.8in]{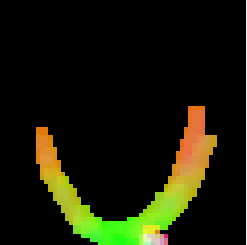}}
\end{center}
  \caption{Grasp representation. (a) A grasp candidate generated from partial point cloud data. (b) Local voxel grid frame. (c-e) Examples of grasp images used as input to the classifier.}
  \label{fig:representation1}
\end{figure}

We represent a grasp candidate to the classifier in terms of the geometry of the observed surfaces and unobserved volumes contained within a region, $R \subset \mathbb{R}^3$, swept out by the parallel jaw fingers as they close. Since the fingers are modeled as rectangles that move toward or away from each other, this region is a rectangular cuboid. Currently, we ignore RGB information. Our representation is illustrated in Figure~\ref{fig:representation1}. Figure~\ref{fig:representation1}(a) shows a grasp candidate generated with respect to partial point cloud data (from the BigBird dataset~\cite{Singh2014}). Figure~\ref{fig:representation1}(b) shows two sets of points in $R$. One set of points, shown in magenta, shows points in the cloud contained within $R$. The other set of points, shown in blue, are sampled from the portion of $R$ that is unobserved, {\em i.e.}, that is occluded from view by every sensor.

In the following, we will assume that the cuboid region $R$ is scaled to the unit cube and the points contained within it are voxelized into a $60 \times 60 \times 60$ grid. For every triple, $(x,y,z) \in [1,60] \times [1,60] \times [1,60]$, $V(x,y,z) \in \{0,1\}$ denotes whether the corresponding voxel is occupied and $U(x,y,z) \in \{0,1\}$ denotes whether the corresponding voxel has been observed. We will further assume that each occupied voxel in $R$ is associated with a unit, outward-pointing surface normal vector, $\hat{n}(x,y,z) \in S^2$, that denotes the orientation of the object surface at that point. All of the above information can be calculated either from the point cloud with associated viewpoints or from a TSDF.

Ideally, we would represent to the classifier the 3D geometry of the object surface contained between the gripper fingers. Unfortunately, this would require passing a large number of voxels to the CNN. Instead, we project the voxels onto  planes orthogonal to the standard basis axes of the voxel grid and pass these to the CNN as input. The directions from which we take these projections are illustrated by the arrows in Figure~\ref{fig:representation1}(b). The red arrow points along the hand approach vector, denoted by the $x$ coordinate in the voxel grid $V$. The green arrow points along a vector parallel to the axis of major principle curvature of the object surface (we will call this the ``curvature axis''), denoted by the $y$ coordinate in $V$. The blue arrow points along an orthogonal direction that views the grasp from the side, denoted by the $z$ coordinate.

For each of these three projections, we will calculate three images: an averaged heightmap of the occupied points, $I_o$, an averaged heightmap of the unobserved region, $I_u$, and averaged surface normals, $I_n$. For example, to project onto the plane constituted by the hand and curvature axes, {\em i.e.}, the (x,y) plane, these maps are calculated as follows:
\[
I_o(x,y) = \frac{\sum_{z \in [1,60]} z V(x,y,z)}{\sum_{z \in [1,60]} 
V(x,y,z)}
\]
\[
I_u(x,y) = \frac{\sum_{z \in [1,60]} z U(x,y,z)}{\sum_{z \in [1,60]} 
U(x,y,z)}
\]
\[
I_n(x,y) = \frac{| \sum_{z \in [1,60]} \hat{n}(x,y,z) V(x,y,z) |}{\sum_{z \in [1,60]} 
V(x,y,z)}
\]
The first two images, $I_o$ and $I_u$, are $60 \times 60$ images. The last image, $I_n(x,y)$, is a $60 \times 60 \times 3$ image where the three dimensions of the normal vector are interpreted as three channels in the image. All together, we have five channels of information for each of the three projections, for a total of 15 channels total.

\subsection{Generating training labels}
\label{sect:labeling}

\begin{wrapfigure}[14]{r}{0.2\textwidth}
\begin{center}
  \includegraphics[height=1.1in]{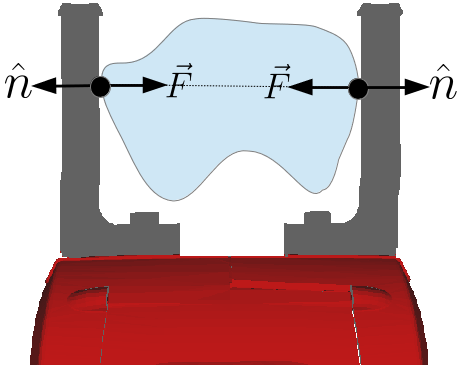}
\end{center}
  \caption{A frictionless antipodal grasp, with contact surface normals anti-parallel to each other.}
  \label{fig:antipodal}
\end{wrapfigure}

In order to classify grasps accurately in Step 2 of Algorithm~\ref{alg:1}, we need to be able to generate a large amount of training data to train the classifier. In particular, we need training data of the following form: a set of exemplars where each exemplar pairs a representation of the geometry or appearance of the local object surface relative to a 6-DOF hand pose with a label indicating whether or not a grasp exists at that hand pose. We generate the training data automatically as follows. First, we use the grasp candidate sampling method described in Section~\ref{sect:gpd} to generate a large set of grasp candidates. Then, we evaluate whether a frictionless antipodal grasp would be formed if the fingers were to close on the mesh from the hand pose associated with the grasp candidate. In the context of a two-finger parallel jaw gripper, a frictionless antipodal grasp means that each of the object surface normals at contact is anti-parallel with the direction in which the contacting finger closes and co-linear with the line connecting the contacts~\cite{Murray1994} (see Figure~\ref {fig:antipodal}). This is a particularly strong condition: a frictionless antipodal grasp is in force closure for any positive coefficient of friction. 


In order to accomplish the above, we require a set of object meshes where each mesh is registered to one or more point clouds (or TSDFs) taken from different viewpoints. This raw data can be obtained using online CAD models such as those contained in 3DNET~\cite{Wohlkinger2012} and simulating the depth images, or it can come from a dataset like BigBird that pairs real depth images with a reconstructed mesh of the object. Unfortunately, determining whether a grasp candidate is 
a frictionless antipodal configuration can be tricky when the object meshes are noisy. (For example, the object meshes in the BigBird dataset are noisy because they are reconstructed from actual sensor data.) We address this by ``softening'' the frictionless antipodal condition described above slightly. In particular, we assume that each vertex in the mesh is subject to a small amount of position error (1mm in our experiments) and evaluate whether an antipodal grasp could exist under any perturbation of the vertices. This reduces to identifying small contact regions in each finger where contact might be established and evaluating whether the frictionless antipodal condition described above holds for any pair of contacts in these regions. 

\section{Evaluation}

Achieving high-precision grasp detection is the key challenge in grasp pose detection. Here, we evaluate the effects of various design choices.

\subsection{Measuring recall-at-high-precision for grasp pose detection}

\begin{figure}[b]
\begin{center}  
  \subfigure[]{\includegraphics[height=1.45in]{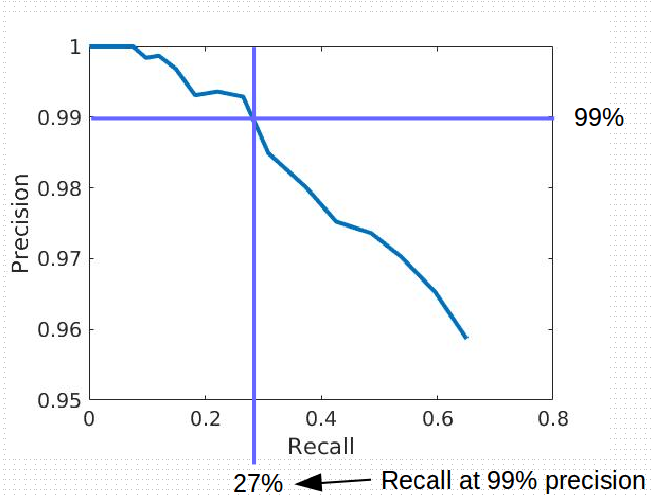}}
  \subfigure[]{\includegraphics[height=1.45in]{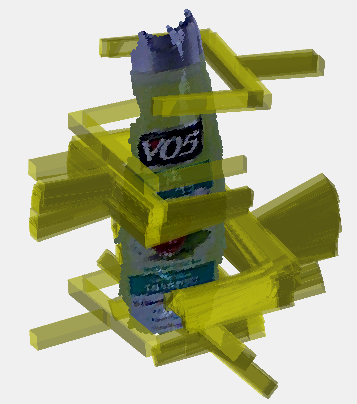}} 
\end{center}
  \caption{Example of recalling grasps at high precision. (a) Precision-recall curve. (b) Grasps recalled at 99\% precision.}
  \label{fig:recall_at_precision}
\end{figure}

Typically, classification performance is measured in terms of accuracy -- the proportion of predictions made by the classifier that match ground truth. Most grasp pose detection systems described in the literature achieve something between 75\% and 95\% grasp classification accuracy~\cite{Fischinger2012, Pinto2016, Lenz2015, Herzog2012, Kappler2015}. Unfortunately, this accuracy number alone does not give us a good indication of whether the resulting grasp pose detection system will have a high grasp success rate. The key question is whether a particular grasp pose detection system can detect grasps with high precision. Precision is the proportion of all positives found by the classifier that are true positives. In grasp pose detection, the cost of a false positive is high because it can cause a grasp attempt to fail. As a result, we want to travel along the precision-recall curve and reach a point with very high precision ({\em i.e.} very few false positives). This amounts to adjusting the classifier acceptance threshold. Setting the threshold very high will result in a high precision, but it will reduce recall -- the proportion of all true positives found by the classifier. Therefore, a key metric for grasp pose detection is recall-at-high-precision. Given a specification that the system must find grasps with a certain minimum precision (say 99\%), what recall can be achieved? This is illustrated in Figure~\ref{fig:recall_at_precision}(a). For a particular shampoo bottle instance, we can recall 27\% of the grasps at 99\% precision. The key insight is that since grasp pose detection systems can detect hundres of grasps for a single object, we don't need to recall {\em all} of the grasps in order to have lots of choices about which grasp to execute. This is illustrated in Figure~\ref {fig:recall_at_precision}(b). Although we are only detecting 27\% of all true positives, there are still plenty of of alternatives.

\subsection{Comparison between different representations}
\label{sect:comparison}

Several representations proposed in the literature can be viewed in terms of the three projections described in Section~\ref{sect:representation}. We compare these representations by evaluating the accuracy with which they can predict grasps.

The dataset we use for this purpose is created based on the point clouds and object meshes in the BigBird dataset. We identify 55 objects (26 box-type objects, 16 cylindrical-type objects, and 13 other objects) out of the 125 objects in the BigBird dataset for which: 1) a complete mesh exists for the object in the dataset; 2) the object can be grasped by a parallel jaw gripper that can open by at most 10cm. From this, we obtain a dataset of 216k grasp candidates evenly balanced between positive and negative exemplars with a 185k/31k train/test split over views (for each object, the test set does not contain any exemplars derived from a view that is present in the training set).

The dataset is created by generating grasp candidates from partial-view point clouds. As in all of our experiments, these partial-view point clouds are created under the assumption that every object is seen by two robot sensors that see the object from viewpoints 53 degrees apart. This stereo-sensor configuration is easy to configure in practice and it reflects the configuration of our robot in the lab. We simulate this configuration by registering together appropriate views from the BigBird dataset. The grasp candidates are labeled by checking whether a frictionless antipodal grasp exists with respect to the full object mesh as described in Section~\ref{sect:labeling}.

\begin{figure}[h]
\begin{center}  
    \includegraphics[height=2.2in,trim={0.4in 0.1in 0.6in 0.5in},clip]{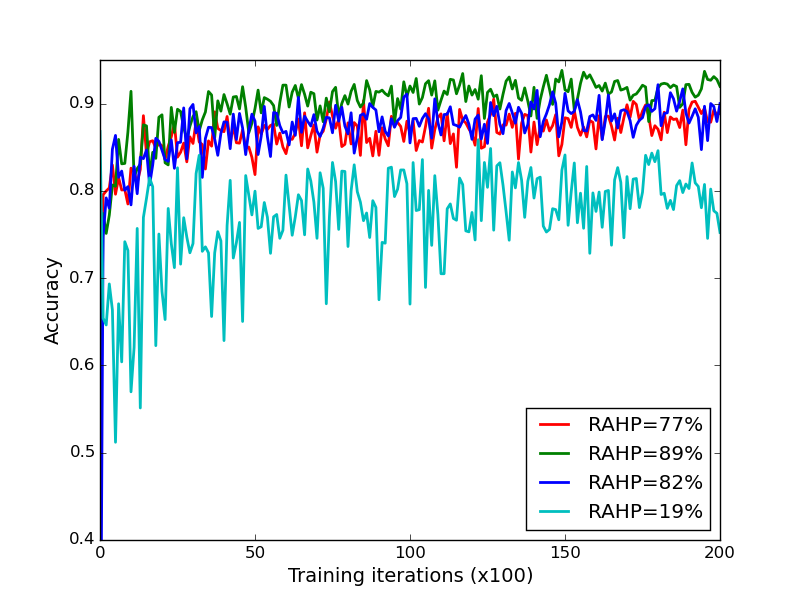}
\end{center}
  \caption{Classification accuracy obtained using different 
  grasp candidate representations. Green: combined 15-channel 
  representation. Blue: same as green but without the occlusion 
  channels. Red: the representation used in our prior work~\cite{tenPas2015}. 
  Cyan: the representation used in both Kappler {\em et al.}~\cite{Kappler2015} and 
  Herzog {\em et al.}~\cite{Herzog2012}. The legend shows the 
  recall-at-high-precision (RAHP) metric for each of these representations 
  for 99\% precision.}
  \label{fig:comparisonOfRepresentations}
\end{figure}

The accuracy of our full 15-channel representation as a function of training iteration (in 100s) is shown in green in Figure~\ref {fig:comparisonOfRepresentations}. Since this train/test split is over views, this result describes the accuracy that would be achieved if we knew we would be given one of the 55 BigBird objects, but did not know in advance which one we would get. We compare the full 15-channel accuracy to the accuracy that can be obtained without using any of the occlusion channels (without $I_u$, a 12-channel representation), shown in blue in Figure~\ref{fig:comparisonOfRepresentations}. Notice that we gain approximately an additional 2\% accuracy by using this information. This is an important piece of information because the occlusion information is difficult and time-consuming to calculate: it requires a TSDF or precise knowledge of the viewpoints and it involves evaluating a relatively large subset of the volume contained between the fingers. The recall-at-high-precision (99\% precision) for the 15-channel and the 12-channel representation is 89\% and 82\%, respectively.

We compare the above to two additional representations to the three-channel representation used in our prior work~\cite{tenPas2015} (shown in red in Figure~\ref{fig:comparisonOfRepresentations}). This representation is comprised of the three-channel $I_n$ image projected along the curvature axis. It is suprising that this representation performs just about as well as the 12-channel, without-occlusion representation above even though it only contains three channels of information. This suggests that beyond the curvature axis projection, the two additional projections do not help much.

We also compare with the representations used in both Kappler {\em et al.} and Herzog {\em et al.} (shown in cyan in Figure~\ref{fig:comparisonOfRepresentations}). That representation is comprised of three channels of information projected along the hand approach axis. One channel is $I_o$. The second channel is $I_u$. The third channel describes the unoccupied voxels in the space: $I_f = \overline{I_u \cup I_o}$. On average, this representation obtains at least 10\% lower accuracy than the other representations and only a 19\% recall-at-high-precision. This lower performance must be due to either or both of these things: 1) projecting along the axis of the hand approach vector loses more information than other projections; 2) not encoding surface normals loses information.

There are several representations in the literature that use the RGB data as well as depth information from a single depth image produced by a Kinect-like sensor. For example, work from Saxena's group detects grasp points in RGBD images~\cite{Lenz2015}. Similarly, Pinto and Gupta~\cite{Pinto2016} use RGBD information as well. In these cases, since the robot hand typically approaches the object from the viewing direction, the depth channel in the above is roughly equivalent to our $I_o$ channel along the hand approach direction. As a result, if the RGB information were not also present, then these representations would fare no better than the Kappler {\em et al.} representation described above. It is possible that the additional RGB information makes up for the deficit, but this seems unlikely given results reported by Lenz {\em et al.} who obtain only 0.9\% additional accuracy by using RGB in addition to depth and surface normal information~\cite{Lenz2015}.

\subsection{Pretraining on simulated data}

\begin{wrapfigure}[15]{r}{0.27\textwidth}
\begin{center}
  \includegraphics[height=1.4in,trim={0.3in 0.1in 0.6in 0.5in},clip]{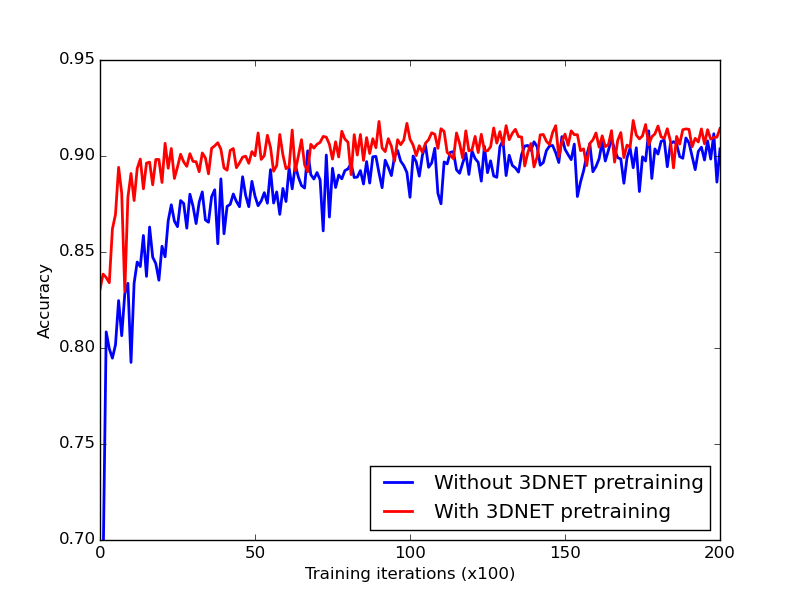}
\end{center}
  \caption{Accuracy with (red) and without (blue) 3DNET pretraining.}
  \label{fig:3dnetpretrain}
\end{wrapfigure}

One way to improve classifier accuracy and precision is to create training data using point clouds or TSDFs created by simulating what a sensor would observe looking at a CAD model. Compared with the amount of real sensor data that is available, there is a huge number of CAD models available online (for example, 3DNET makes available thousands of CAD models from 200 object categories~\cite{Wohlkinger2012}). Ideally, we would train using this simulated data. Unfortunately, there are subtle differences between depth images obtained from real sensors and those obtained in simulation that hurt performance. For example, recall the 31k test set derived from BigBird data described in Section~\ref{sect:comparison}. Our best representation obtained approximately 90\% accuracy over all 55 objects in the BigBird dataset. However, when we train our system for 30000 iterations on 207k exemplars created using on 400 object CAD models taken from 16 categories in 3DNET, we obtain only 83\% accuracy on the same test set. While it is possible that different methods of simulating depth images could improve performance, it is likely that a small difference will persist.

One approach to this problem is to pretrain the CNN learning system using simulated data, but to ``finetune'' it on real data more representative of the problem domain at hand. We evaluated this approach by testing on the 216k BigBird dataset described in Section~\ref{sect:comparison}. We compare the learning curve obtained using the 15-channel representation in Section~\ref{sect:comparison} starting with random network weights with the learning curve obtained using the 3DNET weights as a prior. Figure~\ref{fig:3dnetpretrain} shows the results. The pretrained weights have a strong effect initially: the pretrained network obtains the same accuracy at 4000 iterations as the non-pretrained network obtains after 20000 iterations. However, the importance of the contribution diminishes over time.

\subsection{Using prior knowledge about the object}

A key advantage of grasp pose detection is that it allows us to use varying degrees of prior knowledge to improve detection by adjusting the contents of the training set. If we have no prior knowledge of the object to be grasped, then we should train the grasp detector using data from a large and diverse set of objects. If we know the category of the object to be grasped (for example, if we know the object is box-like), then we should train the grasp detector using training data from only box-like objects. Finally, if we know the exact object geometry, then we should use training data derived only from that particular object. In general, one would expect that the more prior knowledge that is encoded into the network this way, the better our classification accuracy will be. 

\begin{figure}[h]
\begin{center}  
  \includegraphics[height=2.2in]{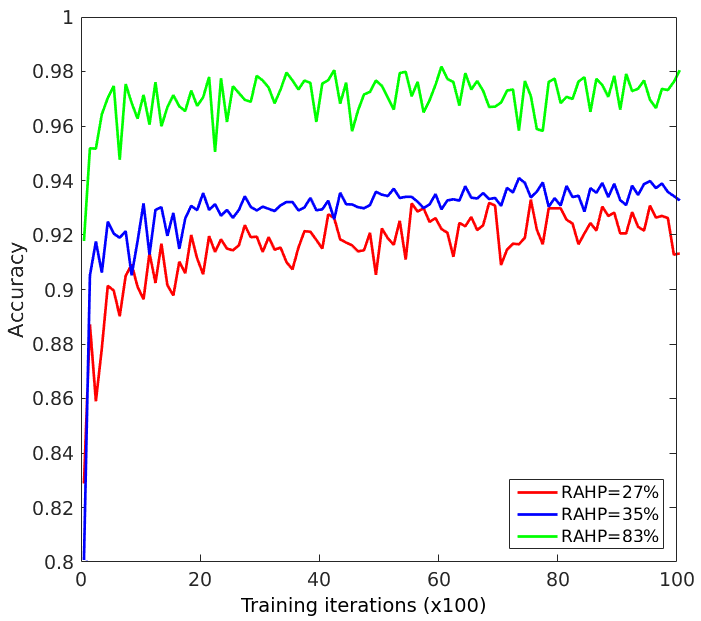}
\end{center}
  \caption{Grasp detection accuracy given no prior knowledge of the 
  object (red); given a category knowledge (blue); given the precise geometry of 
  the object (green). The legend shows the 
  recall-at-high-precision (RAHP) metric for each of these representations 
  for 99\% precision.}
  \label{fig:compareCategoryInfo}
\end{figure}

We performed an experiment using 16 cylindrical-like objects from the BigBird dataset and compared classification accuracy in these three different scenarios. First, we trained a network using training data derived only from the single object in question using a train/test split (45k training and 5k test) on view angle (no one view shared between test and training). This network was trained starting with weights pretrained on the full 55 BigBird object set. Averaged over the 16 cylindrical-like objects, we obtained roughly 97\% classification accuracy and 83\% recall-at-99\%-precision (the green line in Figure~ \ref{fig:compareCategoryInfo}). Second, for each of the 16 cylindrical-like objects, we trained the network using data derived from the other 15 objects (leave-one-object-out). This gave us 150k exemplars for training and 10k for test for each object. This network was pretrained on the 3DNET data. Here we obtained approximately 93.5\% accuracy and 35\% recall-at-99\%-precision (the blue line in Figure~ \ref{fig:compareCategoryInfo}). Finally, for each object, we trained the network using all other objects in the dataset. This gave us 278k training examples and 10k test for each object. Here we obtained approximately 92\% accuracy and 27\% recall-at-99\%-precision (the red line in Figure~\ref{fig:compareCategoryInfo}).

\section{Robot Experiments}

To validate our GPD algorithm, we ran two dense clutter experimental scenarios on a Baxter Research Robot. We used the 3-channel representation (see Section~\ref{sect:comparison}) because it is fast to compute and almost as accurate as the 15-channel representation (see Figure~\ref{fig:comparisonOfRepresentations}). In the first scenario the algorithm received point clouds from two statically mounted depth sensors, and in the second scenario the point cloud information came from a wrist-mounted depth sensor \footnote{Video: \url{https://www.youtube.com/watch?v=p4JXpZVxr48}}.

\subsection{Hardware setup}

\begin{figure}[h]
\begin{center}  
  \includegraphics[height=1.4in,trim={0in 0.2in 0in 0.4in},clip]{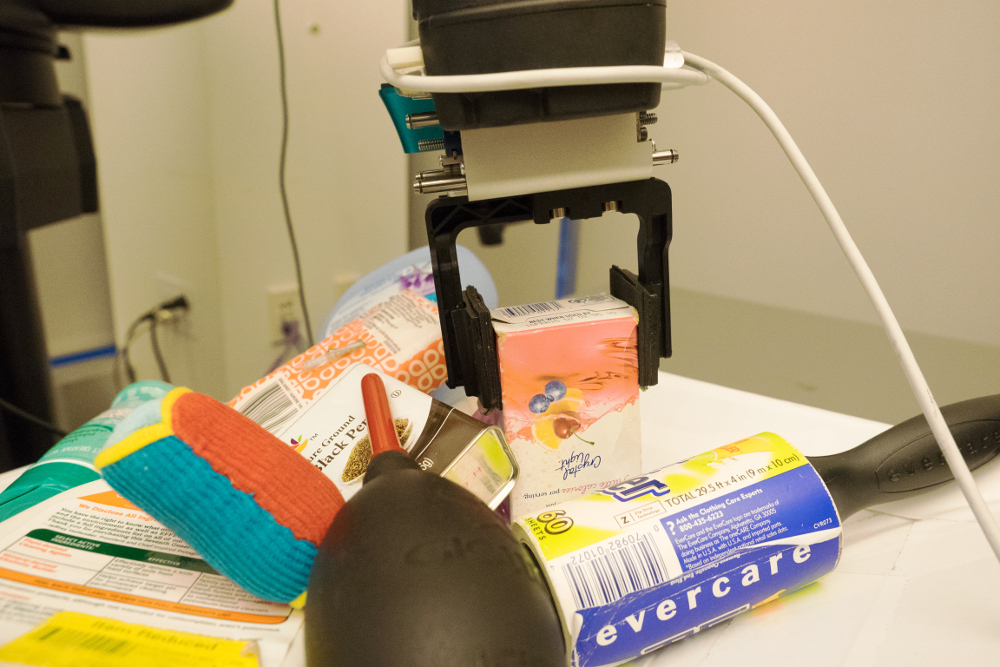}
\end{center}
  \caption{Gripper closing on the first object in the clutter.}
  \label{fig:baxterGrasp}
\end{figure}

We use the right 7-DOF arm of the Baxter Research Robot in the experiments. Our robot hand is the stock Baxter parallel-jaw gripper with the stock, short fingers and square pads. The square pads were modified with a black rubber covering, and rubber-covered pieces of metal were added to the ends (shown in Figure~\ref{fig:baxterGrasp}). The ends bend slightly outward to initially widen the bite which helped with minor, sub-centimeter kinematic or point cloud registration errors. This gripper is restricted to a 3 to 7cm width. Each object in the test set was selected given this restriction. We mounted two Asus Xtion Pro depth sensors to Baxter's waist and an Occipital Structure sensor to the robot's wrist.

We used two computer systems in the experiments. Each system consisted of a 3.5 GHz Intel Corei7-4770K CPU (four physical cores), 32 GB of system memory, and an Nvidia GeForce GTX 660 graphics card. One system was used to run our GPD algorithm, and we used InfiniTAM~\cite{Kahler2015} on the other system to obtain a TSDF volume from the wrist-mounted sensor while moving the robot arm. Communication between the robot and the two PCs was handled by the robot operating system (ROS). TrajOpt~\cite{Schulman2013} was used for motion planning.

\subsection{Experimental protocol}

\begin{figure}[h]
\begin{center}
  \subfigure[]{\includegraphics[height=1.3in,trim={0.2in 0in 0in 1.0in},clip]{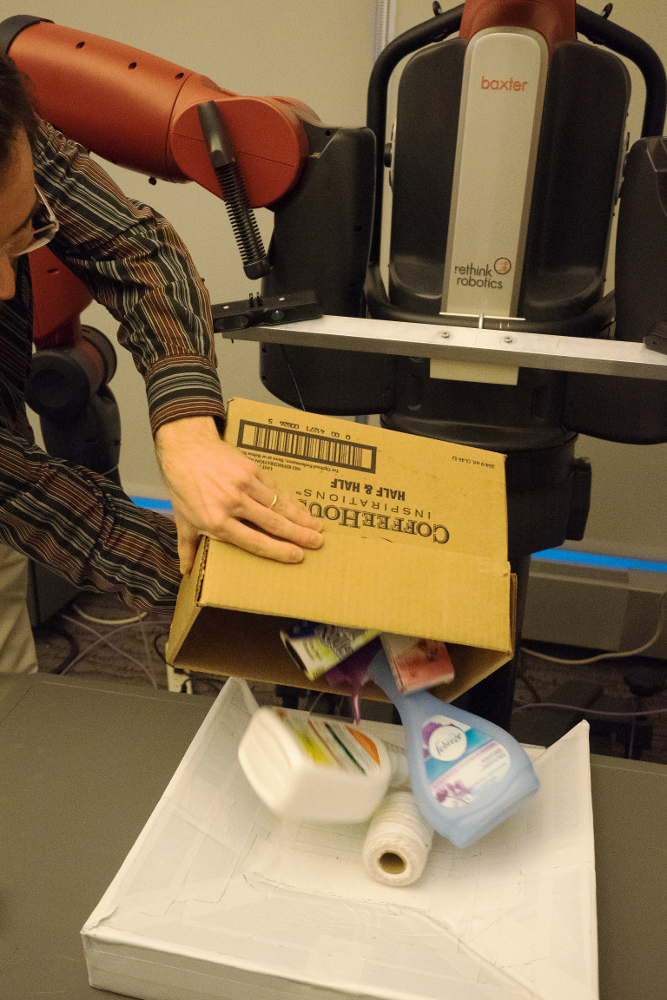}}
  \subfigure[]{\includegraphics[height=1.3in,trim={0.4in 0in 0.4in 0in},clip]{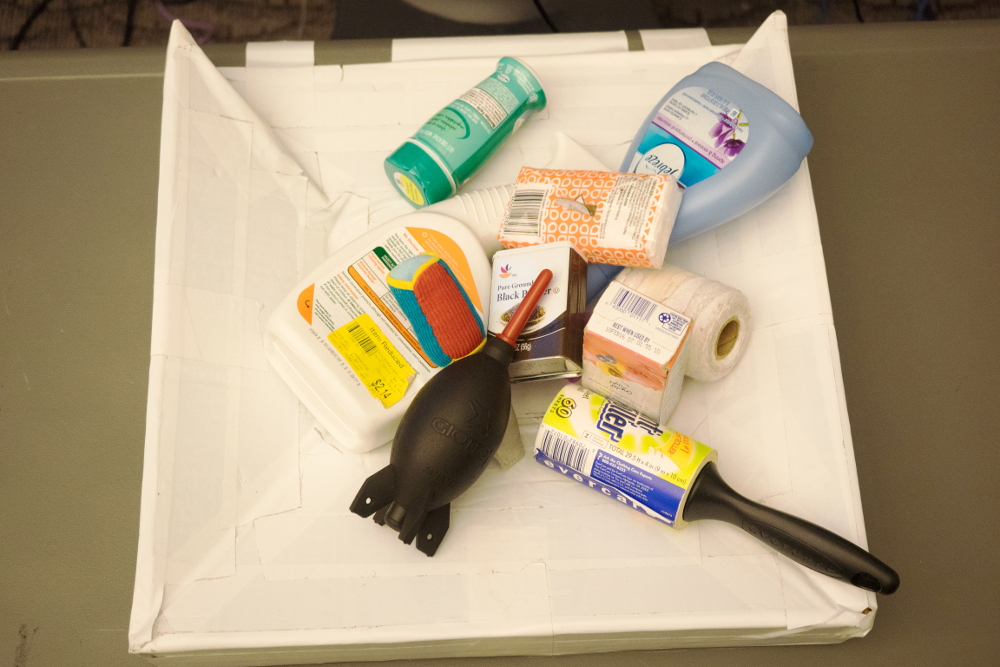}}
\end{center}
  \caption{Preparation of a clutter clearing task for Baxter. (a) Pouring the box contents into the tray. (b) Tray contents immediately after pouring.}
  \label{fig:baxterDumpTray}
\end{figure}

We ran 45 clutter experiments, of which 15 rounds were tested with a static placement of two Kinect sensors (we call this the ``passive scenario''), and of which 30 rounds were tested with the wrist-mounted sensor while streaming images (we call this the ``active scenario''). The protocol for each round is outlined in Table~\ref{table:experimentProtocol}. First, 10 objects were selected uniformly at random from a set of 27. These 10 objects were used for both the passive round and two active rounds. The 27 objects are the same common household items that we used in our prior work~\cite{tenPas2015}, none of which are in the training set. All of the objects are lighter than 500g and have at least one side that fits within the gripper. Figure~\ref{fig:baxterDumpTray} shows a person preparing one round of the experiment.

\begin{table}[h!]
\centering
\begin{tabular}{|l | l|} 
 \hline
 1. & Randomly select 10 objects from the object set.\\
 \hline
 2. & Place objects into a box.\\
 \hline
 3. & Shake box until sufficiently mixed.\\
 \hline
 4. & Pour box contents into tray in front of robot.\\
 \hline
 5. & Run clutter removal algorithm (see Section~\ref{sect:clutterRemovalAlgorithm}).\\
 \hline
 6. & Terminate once any of these events occur:\\
 & i) No objects remain in tray.\\
 & ii) No grasp hypotheses were found after 3 attempts.\\
 & iii) The same failure occurs on the same object 3 times.\\
 \hline
\end{tabular}
\caption{Clutter-removal experiment protocol for one round.}
\label{table:experimentProtocol}
\end{table}

The next steps were to place the 10 objects in a box and to shake the box to mix the objects. We then poured the contents of the box into a tray placed in front of the robot on a table. Two objects (the sandcastle and the lobster) could not be grasped from an upside down configuration. If either of these objects fell into this configuration, then we manually removed the object, turned it right side up, and placed it back on the pile. The experiment proceeds until either there are no objects remaining in the tray, until the GPD algorithm has run three times and no grasp hypotheses were found, or until the same failure occurs on the same object three times in a row. The latter case only occurred once in our experiments (in the active scenario), where the front of the vacuum attachment was not grasped in a stable way.

\subsection{Clutter removal algorithm}
\label{sect:clutterRemovalAlgorithm}

Our algorithm for clearing the clutter works as follows. For the passive scenario, each Kinect sensor takes an image, and both images are transformed into the robot's base frame, combined, and voxelized. The positions of the sensors themselves are hard-coded into the algorithm and used for the surface normal calculation. For the active scenario, we use InfiniTAM to integrate a TSDF volume of the cluttered scene~\cite{Kahler2015} as the arm travels about 46cm along a fixed trajectory. We used TrajOpt to generate this trajectory offline and reused it in all experiments for the active scenario. The trajectory was constrained to align the line-of-sight axis of the sensor towards a point centered in the cluttered tray and to keep a minimum distance of 40cm between this fixed point and the origin of the sensor. (The minimum range of the sensor is between 35 to 40cm.)


The next step is to run the GPD algorithm. The GPD algorithm typically returns hundreds of grasp candidates in about 3s. In order to select the grasp that is expected to be most likely to succeed, we first use IKFast~\cite{Diankov2010} to find collision-free IK solutions for each grasp and remove those that have no solution. We use OpenRAVE \cite{Diankov2010} for collision checking, and we model the point cloud obstacle as 2cm cubes centered at points in a voxelized cloud. Our next step is to prune grasps that are very close to the joint limits of the robot arm or require a width that is less than 3cm or more than 7cm. For each remaining grasp, we modify its position by the average position of all other grasps that are similar in position and orientation. This is important because it makes it more likely that the grasp will succeed if the gripper is slightly off due to kinematic or point cloud registration errors.

After filtering grasps, we compute a utility function that takes the height of the grasp position, the required gripper width, the angle between the grasp approach vector and the robot base vertical axis, and the c-space distance the arm needs to travel to reach the grasp from a fixed, nominal configuration into account. We have empirically found it is a good strategy to grasp the highest object and approach from the top because of potential collisions with other objects in the tray. Additionally, the distance the arm travels is important because a smaller distance is less likely to put the arm in a position where the kinematic readings are significantly off, and it also takes less time.

\subsection{Results}
\label{sect:robot_results}

The results of our experiments are presented in Table~\ref{table:results_robot}. For the 15 rounds of experiments with the passive, statically placed sensors, we observed an 84\% grasp success rate (22 grasp failures out of 138 grasp attempts). Of the 22 failures, 5 were due to point cloud registration or kinematic errors of the robot, 13 were due to perception errors caused by our GPD algorithm, 2 were due to a collision of the fingers with the object before the grasp, and 2 were due to the object dropping out after an initially successful grasp. In this scenario, 77\% of the objects placed in front of the robot were cleared. The others were either knocked out of the tray, pushed too far forward to be seen by the sensors, or grouped too close together for a finger-clear grasp to be found.

For the 30 rounds of experiments with the active, wrist-mounted sensor, we observed a 93\% grasp success rate (20 grasp failures out of 288 grasp attempts). Out of the 20 failures, 5 were due to point cloud registration or kinematic errors of the robot, 9 were due to perception errors caused by our GPD algorithm, 4 were due to a collision of the fingers with the object before the grasp, and 2 were due to the object dropping out after an initially successful grasp. In this scenario, 90\% of the objects were cleared from the tray. This improvement is partially due to the front of the tray being more visible to the wrist-mounted sensor, but otherwise some objects were still left behind for the same reasons as in the passive scenario.

The 9\% grasp success rate improvement in the active scenario compared to the passive scenario is due to to having a better and more complete view of the objects in the tray. If the objects are not seen at all, a collision can occur which may cause the target object to move out of the way before the gripper closes. Also, the more partial the point cloud of the objects, the more difficult it is for GPD to correctly classify the grasps. 

\begin{table}[h!]
\centering
\begin{tabular}{|l|r|r|}
  \hline
  & Passive scenario & Active scenario\\
  \hline 
  Grasp success rate & 84\% & 93\%\\
  \hline
  No. of objects & 150 & 300 \\
  \hline
  No. of objects removed & 116 (77\%) & 269 (90\%) \\
  \hline
  No. of grasp attempts & 138 & 288 \\
  \hline
  No. of grasp failures & 22 & 20\\
 \hline
\end{tabular}
\caption{Results of the clutter-removal experiments.}
\label{table:results_robot}
\end{table}

\section{Conclusion}

In this paper, we have built up on previous work on grasp pose detection in point clouds that first creates a large set of grasp hypotheses and then classifies them as a good or a bad grasps. We proposed three ways of improving the classification step. First, we introduced a new 15-channel grasp candidate representation as input to train a CNN. We compared this representation to existing representations from the literature and to our own representation from previous work. We find that the more information our representation encodes, the better the classifier performs. However, our 15-channel representation performs just slightly better than our 12-channel representation; although it is much more expensive to compute. Second, we found that a network pretrained on simulated depth data obtained from online data sets reduces the number of iterations required to achieve high classification accuracy compared to an untrained network. Third, using prior instance or category information about the object to be grasped improves classification performance. Compared to our previous work, our robot experiments in this paper show an increase of the average success rate from 73\% to 93\% for grasping novel objects in dense clutter. This result emphasizes that the ways we proposed to enhance the classification significantly improve robot grasping in the real world.

GPD algorithms are currently limited by not being able to infer non-geometric object properties such as center of mass, inertia, and weight. Moreover, as these algorithms typically avoid point cloud segmentation, there is no direct way to grasp a specific object of interest and to distinguish between two adjacent objects. In future work, we hope to address these issues.

\section*{Acknowledgements}
This work was supported in part by NSF under Grant No.\ IIS-1427081, NASA under Grant No.\ NNX16AC48A and NNX13AQ85G, and ONR under Grant No.\ N000141410047. Kate Saenko was supported by NSF Award IIS-1212928.

\bibliographystyle{plain}
\bibliography{References.bib}

\end{document}